\documentclass[10pt,twocolumn,letterpaper]{article}

\usepackage[pagenumbers]{cvpr} 
\usepackage{times}
\usepackage{epsfig}

\usepackage{graphicx}
\usepackage{amsmath}
\usepackage{amssymb}
\usepackage{booktabs}

\usepackage{algorithm}
\usepackage{algpseudocode}
\usepackage{array}
\usepackage{colortbl}
\usepackage{multirow}
\usepackage{tabularx}
\usepackage{cite}

\newcommand{\RomNum}[1]{\MakeUppercase{\romannumeral #1}}
\definecolor{darkblue}{rgb}{0,0,0.88}

\usepackage[pagebackref,breaklinks,colorlinks]{hyperref}

\usepackage[capitalize]{cleveref}
\crefname{section}{Sec.}{Secs.}
\Crefname{section}{Section}{Sections}
\Crefname{table}{Table}{Tables}
\crefname{table}{Tab.}{Tabs.}

\def\cvprPaperID{6920} 
\def\confName{CVPR}
\def\confYear{2022}

\begin{document}

\title{Moving Window Regression: A Novel Approach to Ordinal Regression}

\author{Nyeong-Ho Shin\\
Korea University\\
{\tt\small nhshin@mcl.korea.ac.kr}
\and
Seon-Ho Lee\\
Korea University\\
{\tt\small seonholee@mcl.korea.ac.kr}
\and
Chang-Su Kim\\
Korea University\\
{\tt\small changsukim@korea.ac.kr}
}

\maketitle

\begin{abstract}
  A novel ordinal regression algorithm, called moving window regression (MWR), is proposed in this paper. First, we propose the notion of relative rank ($\rho$-rank), which is a new order representation scheme for input and reference instances. Second, we develop global and local relative regressors ($\rho$-regressors) to predict $\rho$-ranks within entire and specific rank ranges, respectively. Third, we refine an initial rank estimate iteratively by selecting two reference instances to form a search window and then estimating the $\rho$-rank within the window. Extensive experiments results show that the proposed algorithm achieves the state-of-the-art performances on various benchmark datasets for facial age estimation and historical color image classification. The codes are available at \url{https://github.com/nhshin-mcl/MWR}.
\end{abstract}
\vspace*{-0.40cm}
\section{Introduction}

Ordinal regression aims to predict the rank of an object instance. It is widely used for computer vision tasks, including facial age estimation \cite{niu2016ordinal} and historical color image (HCI) classification \cite{palermo2012dating}. Thus, various ordinal regression techniques have been developed.

Rank estimation, however, remains challenging because there is no clear distinction between ranks in many cases. For example, in facial age estimation, the aging process~\cite{albert2007age, zebrowitz1997reading}, causing variations in facial shapes, sizes, and texture, has large individual differences due to numerous factors such as genes, diet, and lifestyle, and there are no clear aging characteristics in each age class. To address this issue, extensive researches have been carried out. Recently, Li \etal \cite{li2019bridgenet} used tens of local regressors, each of which learns aging characteristics within a specific age range. Also, Lim \etal \cite{lim2020order} proposed the notion of order learning, and Lee and Kim \cite{lee2021repulsive} improved it based on the order-identity decomposition.

 \begin{figure}[t]
    \centering
   \includegraphics[width=\linewidth]{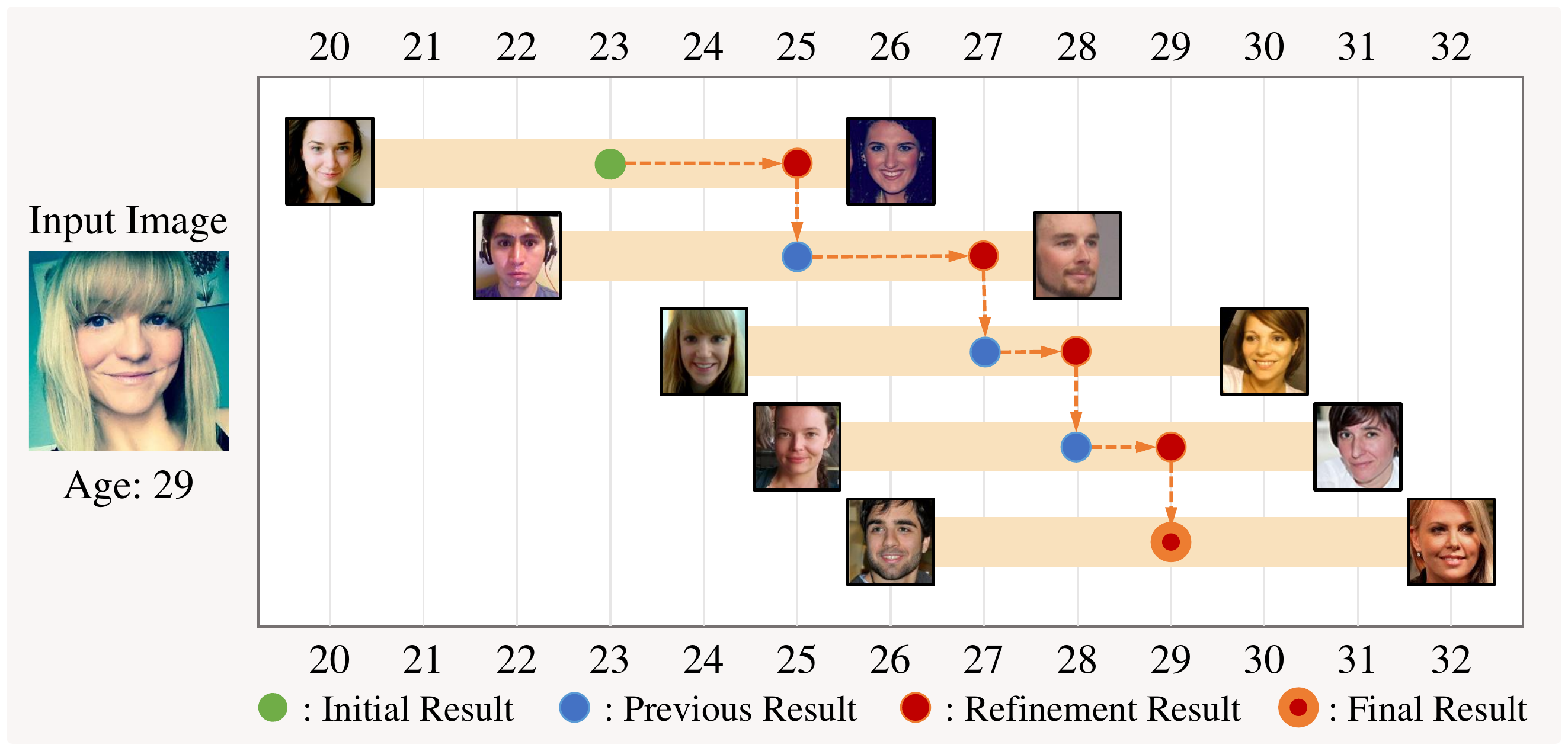}
    \caption{In the MWR algorithm for facial age estimation, an initial rank (age) estimate of the input image is refined iteratively by estimating the $\rho$-rank within a search window (orange bar). The window is bounded by the known ranks of two reference images, and the $\rho$-rank quantifies how much older or younger the input is than the references. Each window is centered around the previous result. It is strongly recommended to watch the accompanying video for an overview of MWR. Also, note that MWR is applicable to general ordinal regression tasks, as well as age estimation.}
    \label{fig:relative_age}
 \end{figure}

To measure the quantity of something, it would be easier and more accurate if some references are available~\cite{blumenthal1977process}. For example, we can estimate the length of an object more accurately if we have a one-meter bar as a reference. In this case, our brains measure the ratio between the lengths of the object and the bar, instead of the absolute length of the object. Similarly, multiple references of different ranks can offer a useful ordinal scale in rank estimation, as a one-meter bar does in the length measurement. Therefore, rather than predicting the absolute rank of an instance directly, we attempt to estimate the relative rank that quantifies how much greater or smaller the instance is than the references.

Based on the observations, we propose a novel ordinal regression algorithm, called moving window regression (MWR), which is illustrated in Figure \ref{fig:relative_age}. First, we propose the notion of relative rank ($\rho$-rank), which quantifies the ordinal relations among the ranks of input and reference instances: the $\rho$-rank measures how much greater the input is than the first reference and how much smaller it is than the second one. Second, to estimate the $\rho$-rank, we develop a relative regressor ($\rho$-regressor), composed of an encoder and a regression module. Third, we propose the MWR process in  Figure \ref{fig:relative_age}. It obtains an initial rank estimate of an input instance based on the nearest neighbor (NN) criterion. Then, it refines the estimate iteratively by selecting two reference instances to form a search window and estimating the $\rho$-rank within the search window. Here, each window is centered around the previous estimate. This is iterated until the convergence. Last, to cope with diverse characteristics in different rank groups, we develop local $\rho$-regressors for those groups, as well as the global $\rho$-regressor for the entire rank range.

This work has the following major contributions:
\begin{itemize}
\itemsep0em
\item We first propose the notion of $\rho$-rank and design $\rho$-regressors for estimating $\rho$-ranks.
\item We develop the novel MWR algorithm for accurate rank estimation, which iteratively predicts the $\rho$-rank within a moving search window.
\item MWR provides the state-of-the-art performances on various benchmark datasets for facial age estimation and HCI classification. Especially, for facial age estimation, MWR performs the best in 17 out of 19 benchmark tests.
\end{itemize}

\section{Related Work}

\subsection{Ordinal Regression}

In ordinal regression, the rank of an object is estimated. In many cases, ordinal regression is converted into multiple binary classification problems \cite{frank2001simple,li2007ordinal}. Liu \etal \cite{liu2017deep} developed a deep ordinal regressor for small datasets. They also proposed another ordinal regressor \cite{liu2018constrained} to adopt the multi-class classification loss additionally. Fu \etal~\cite{fu2018depth} addressed monocular depth estimation based on ordinal regression. D\'{i}az \etal \cite{diaz2019soft} used a soft ordinal label to train an ordinal regressor. Li \etal \cite{li2021ordinal} developed a probabilistic embedding method for ordinal regression.

Order learning~\cite{lim2020order} learns ordering relations between instances. By comparing a test instance with references with known ranks, it can estimate the rank of the instance. In other words, it can perform ordinal regression. Lee and Kim~\cite{lee2021repulsive} proposed the deep repulsive clustering and the order-identity decomposition for effective order learning. Their rank estimation algorithm was applied successfully to facial age estimation. The rationale behind order learning is that relative rank comparison of two instances is easier than absolute rank estimation of each instance. In this paper, we adopt this idea and propose a novel ordinal regression algorithm through relative comparisons. However, whereas order learning is based on ternary classification, the proposed MWR yields a continuous regression score, called $\rho$-rank, indicating the relative rank of an input instance between two references. Thus, unlike order learning, for example, in facial age estimation, given two references of ages 20 and 30, MWR can directly regress a continuous age of a test instance, which is older than 20 and younger than 30.

\subsection{Applications}
Facial age estimation and HCI classification are the most representative vision applications, on which new ordinal regression methods are tested and compared. Let us briefly review conventional ordinal regression methods for these two tasks.

\vspace*{0.15cm}
\noindent\textbf{Facial age estimation:} It is one of the most popular and the most challenging ordinal regression tasks, whose goal is to predict people's ages using their facial images. Various age estimators have been developed, which can be grouped into four categories: classification \cite{rothe2018imdb, tan2018efficient}, regression \cite{shen2018forests,li2019bridgenet}, ranking \cite{niu2016ordinal, chen2017ranking}, and distribution learning \cite{pan2018mean, wen2020adaptive} methods.

OR-CNN~\cite{niu2016ordinal} and Ranking-CNN~\cite{chen2017ranking} are the ranking (or ordinal regression) methods, which regard ages as ranks. To estimate a person's age, they dichotomize whether the person is older than each age or not. By combining these binary classification results, the age can be estimated \cite{frank2001simple,li2007ordinal}. While OR-CNN uses a common feature for all binary classifiers, Ranking-CNN uses different features for different classifiers. However, based on binary classifiers, these methods disregard the continuity of the aging process.

\vspace*{0.15cm}
\noindent\textbf{HCI classification:} It aims at estimating the shooting decade of a photograph. Palermo \etal \cite{palermo2012dating} introduced this task and perform the classification by exploiting different characteristics of the color imaging process in each decade. Martin \etal \cite{martin2014dating} developed a binary classifier to predict whether a photograph was taken earlier or later than each decade and combined multiple classification results to yield a final estimate. Liu \etal \cite{liu2018constrained} used an additional classification loss for training their ordinal regressor. Li \etal \cite{li2021ordinal} proposed a new feature embedding scheme for ordinal regression.



\section{Proposed Algorithm}
We propose the MWR algorithm for ordinal regression. First, we introduce the notion of $\rho$-rank. Then, we develop two types of relative regression networks: global and local $\rho$-regressors designed for entire and specific rank ranges, respectively. Using them, we predict the $\rho$-rank of an instance by comparing it with selected reference instances. Lastly, we estimate the absolute rank via an iterative refinement process, called MWR.

We use facial age estimation examples for concrete description of MWR in this section. However, note that MWR can be applied to other ordinal regression tasks as well.

\subsection{$\rho$-Rank}
Given an object instance $x$, we aim at estimating the corresponding rank $\theta(x)$. However, in many cases, rank estimation is challenging. For example, in age estimation, the facial aging process has large individual variations. People of the same age often look quite different from one another, which makes it difficult to train a reliable network for age estimation.

To deal with this issue, we propose the notion of $\rho$-rank. Note that it is easier to tell the age ordering relations among multiple people than to estimate the exact age of each person~\cite{chang2010ranking,zhang2017comparisons,lim2020order}. Hence, instead of estimating the absolute rank $\theta(x)$ of input $x$, we predict its $\rho$-rank
\begin{equation}\label{eq:relative_age}
    \rho(x, y_1, y_2) = \frac{\theta(x) - \mu(y_1, y_2)}{\tau(y_1, y_2)}
\end{equation}
where $y_1$ and $y_2$ are two references with $\theta(y_1) < \theta(y_2)$. Also, $\mu(y_1, y_2) = \frac{1}{2}(\theta(y_1) + \theta(y_2))$ is the average rank of the two references, and $\tau(y_1, y_2) = \frac{1}{2}(\theta(y_2) - \theta(y_1))$ is half of the rank difference between them.

Suppose that $\theta(y_1) \leq \theta(x) \leq \theta(y_2)$. Then, the $\rho$-rank in \eqref{eq:relative_age} has the following properties. First, $\rho \in [-1, 1]$. Second, the sign of $\rho$ represents whether the rank of input $x$ is closer to that of $y_1$ or $y_2$. It is positive when $x$ is closer to $y_2$ and negative otherwise. Third, the absolute value of $\rho$ quantifies the closeness of $x$ to $y_1$ or $y_2$. For instance, $|\rho| = 1$ only if $\theta(x)=\theta(y_{1})$ or $\theta(y_{2})$. Finally, the absolute rank $\theta$ can be reconstructed from the $\rho$-rank by \begin{equation}\label{eq:absolute_age}
    \theta(x) = \rho(x, y_1, y_2) \cdot \tau(y_1, y_2) + \mu(y_1, y_2).\
\end{equation}
Hence, we predict the $\rho$-rank and then reconstruct the absolute rank $\theta$ from the predicted $\rho$-rank via \eqref{eq:absolute_age}.

\subsection{$\rho$-Regressor}

\begin{figure}[t]
    \centering
    \includegraphics[width=1\linewidth]{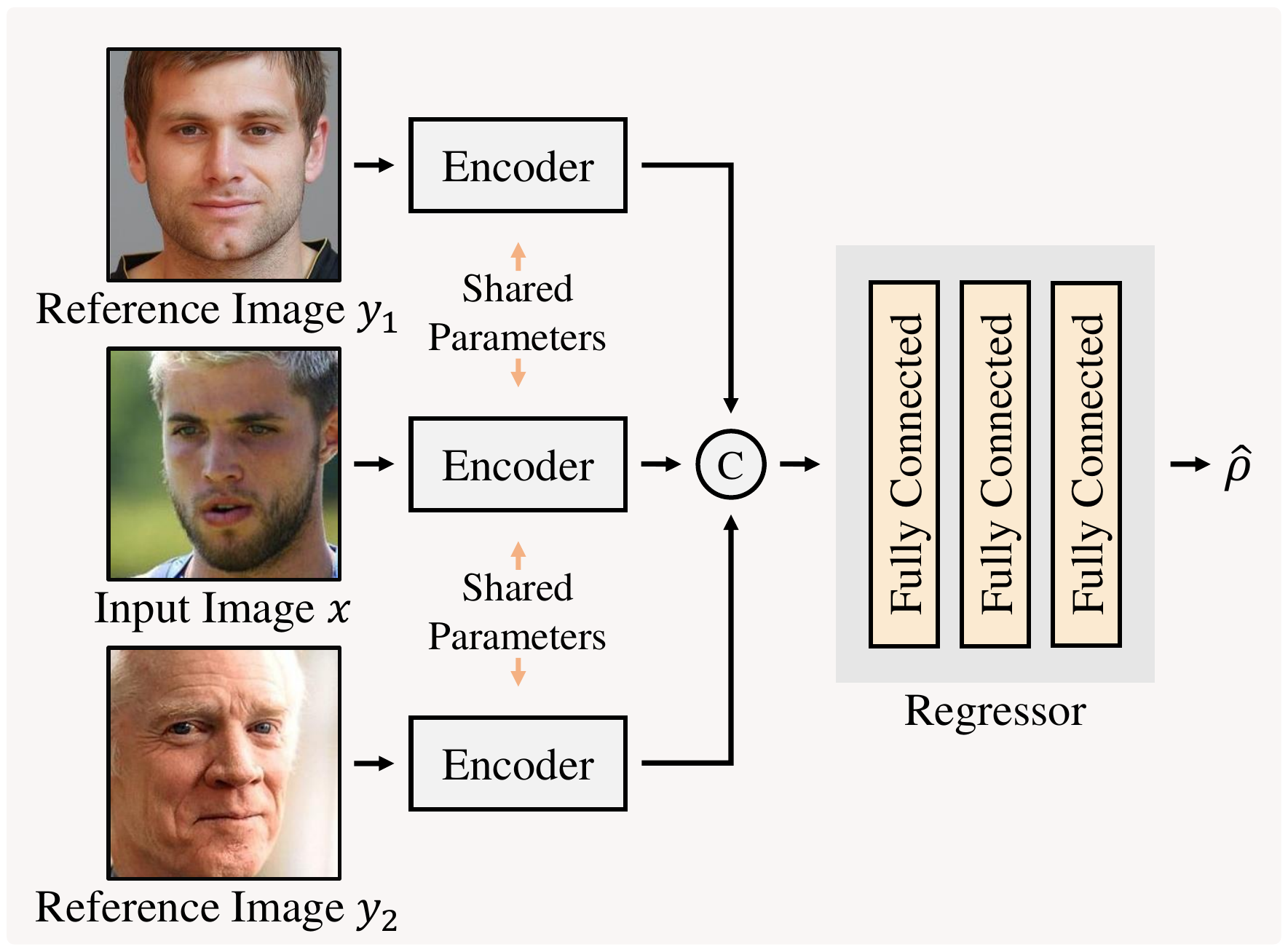}
    \caption{An overview of the $\rho$-regressor. Given an input instance $x$ and two references $y_1$ and $y_2$, the encoder extracts their features separately. After concatenating those features, the regressor yields an estimate of the $\rho$-rank $\hat{\rho}(x, y_1, y_2)$. Here, \textcircled{c} denotes concatenation.}
    \label{fig:Age_Regressor}
\end{figure}

We develop a regressor to predict the $\rho$-rank in \eqref{eq:relative_age}. The proposed $\rho$-regressor in Figure~\ref{fig:Age_Regressor} consists of an encoder $f(\cdot)$ and a regression module $g(\cdot)$.

We adopt VGG16~\cite{simonyan2015vgg} as the encoder. Specifically, we use the output of the last pooling layer in VGG16 as the feature vector $f(x)$. Note that the $\rho$-rank of $x$ is determined in the context of two references $y_1$ and $y_2$. Therefore, the features $f(y_1)$ and $f(y_2)$ are also extracted by the same encoder, as shown in Figure~\ref{fig:Age_Regressor}. Next, the regression module takes the triplet  $(f(x), f(y_1), f(y_2))$ and yields an estimate of the $\rho$-rank in \eqref{eq:relative_age}, which is given by
\begin{equation}
\label{eq:r_estimate}
\hat{\rho}(x, y_1, y_2) = g(f(x), f(y_1), f(y_2)).
\end{equation}
The regression module comprises three fully connected layers: the first two adopt the ReLU~\cite{agarap2018relu} activation, while the last one uses the $\tanh$ activation to yield a value in $[-1, 1]$.

The $\rho$-regressor is end-to-end trained. We form a triplet $(x, y_1, y_2)$ by choosing the references with a fixed
\begin{equation}
\tau=\frac{1}{2}(\theta(y_{2})-\theta(y_{1})).
\label{eq:arithmetic_tau}
\end{equation}
In other words, the rank difference between $y_1$ and $y_2$ is constrained to be a constant. This is to lower the learning difficulty of the $\rho$-regressor. By fixing $\tau$, the $\rho$-regressor needs to consider a much smaller subset of $\{(x, y_1, y_2)\}$ and can achieve more reliable regression. 
Given the triplet $(x, y_1, y_2)$, the $\rho$-regressor obtains the estimate $\hat{\rho}$ in \eqref{eq:r_estimate}. Then, its squared error from the ground-truth $\rho$ in \eqref{eq:relative_age} is defined as the loss, and the $\rho$-regressor is trained to minimize such losses over numerous triplets. When $\theta(x)<\theta(y_1)$ or $\theta(x)>\theta(y_2)$, the ground-truth $\rho$ is set to $-1$ or $1$, respectively.

\begin{figure}[t]
    \centering
    \subfloat[]{\includegraphics[width=\linewidth]{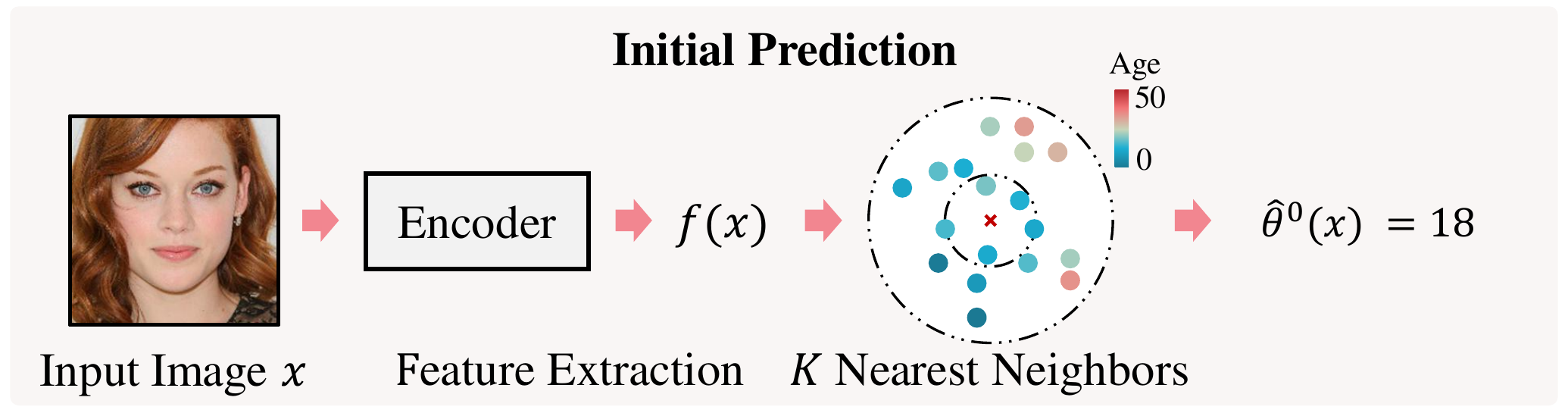}} \\
    \subfloat[]{\includegraphics[width=\linewidth]{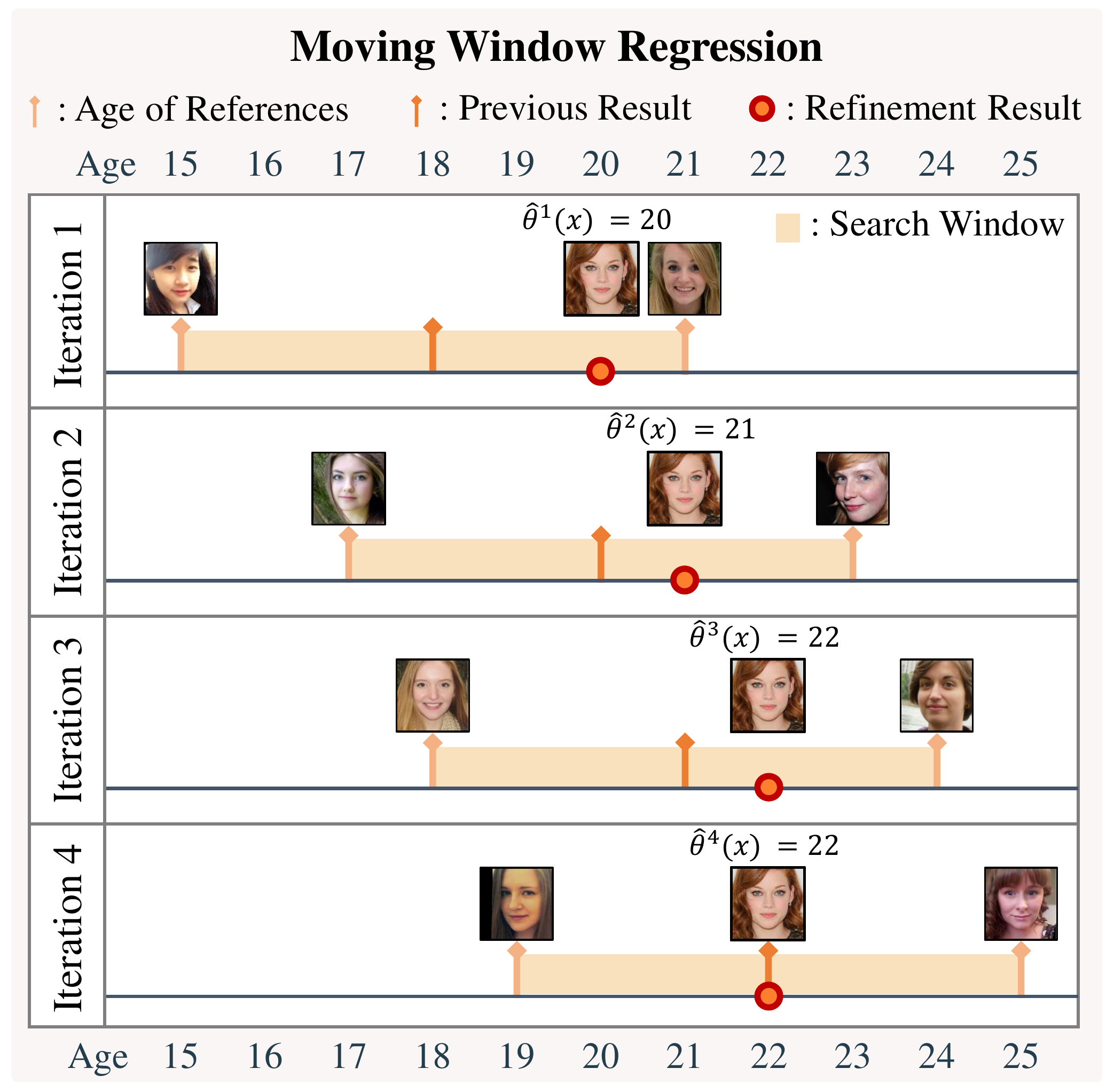}}
    \vspace{-0.2cm}
    \caption{An example of the MWR process in facial age estimation when the ground-truth age of input $x$ is 22 and $\tau$ equals 3: (a) initial prediction and (b) iterative MWR refinement.}
    \label{fig:MWM}
\end{figure}

\subsection{Moving Window Regression}
\label{ssec:MWM}

We estimate the absolute rank $\hat{\theta}(x)$ of an unseen test instance $x$ by moving a window $[\theta(y_1), \theta(y_2)]$ and predicting the $\rho$-rank $\hat{\rho}(x, y_1, y_2)$ in \eqref{eq:r_estimate} using the $\rho$-regressor. This MWR process is performed iteratively.

\begin{figure}[t]
    \centering
    \includegraphics[width=0.95\linewidth]{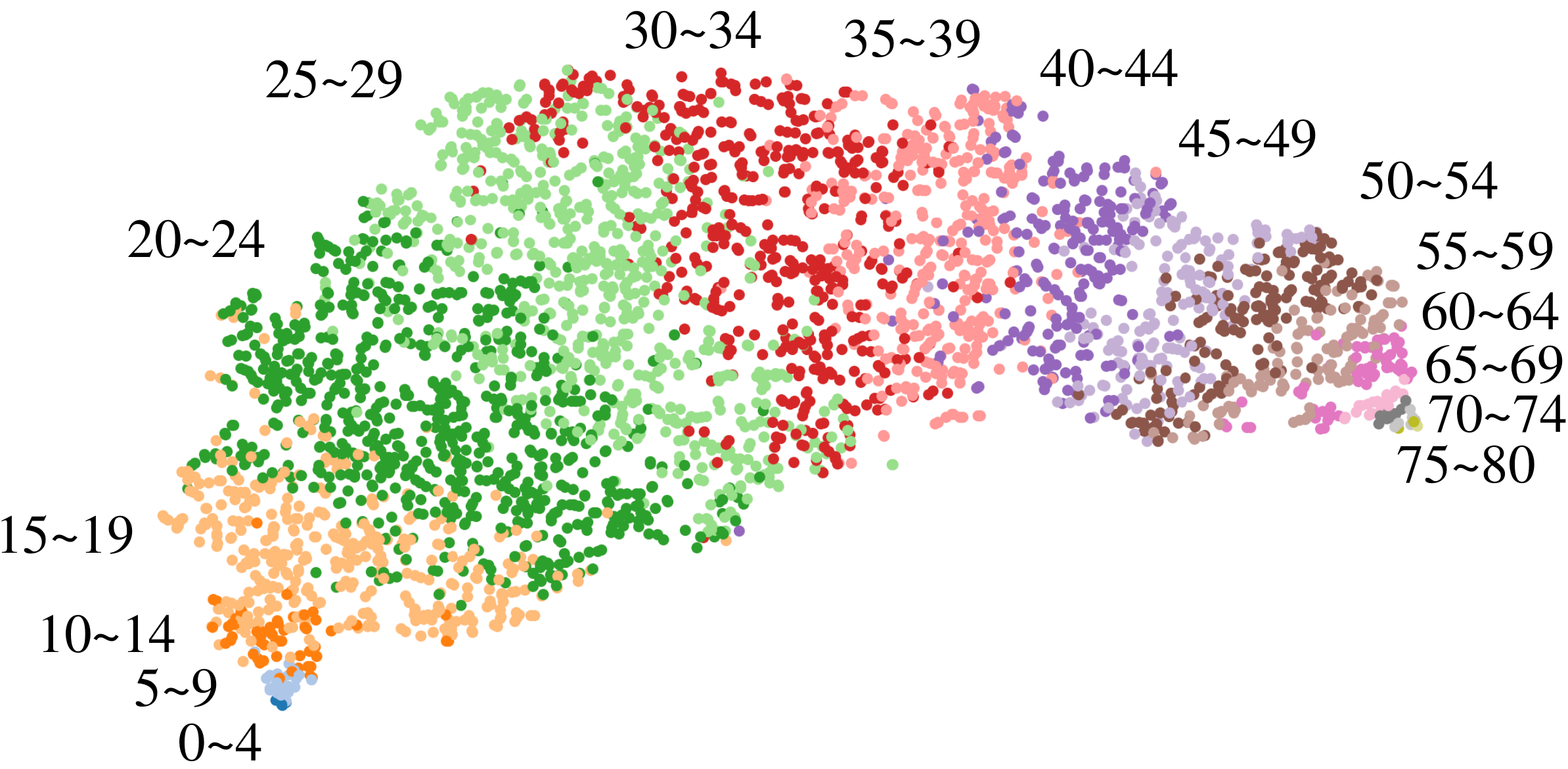}
    \caption{t-SNE visualization \cite{maaten2008tsne} of the feature space of the CLAP2015 dataset \cite{escalera2015chalearn}. Note that feature vectors are aligned roughly according to the ages.}
    \label{fig:tnse-viz}
\end{figure}

First, we obtain an initial estimate $\hat{\theta}^{0}(x)$, where the superscript denotes the iteration index. Figure~\ref{fig:MWM}(a) illustrates this initial estimation. The encoder extracts the feature vector $f(x)$. Then, in the feature space, we find the $K$ NNs to $x$ among all training instances in terms of the Euclidean distances. In this work, $K$ is set to 5. The average rank of these neighbors is rounded to the nearest integer, which becomes the initial estimate $\hat{\theta}^{0}(x)$. This is reasonable since feature vectors are clustered according to the ranks, as visualized in Figure~\ref{fig:tnse-viz}.

Next, at iteration step $t$, we refine $\hat{\theta}^{t-1}(x)$ to $\hat{\theta}^{t}(x)$, as in Figure~\ref{fig:MWM}(b). Among the training instances, we select a pair of references $y_1^t$ and $y_2^t$, whose ranks are $\theta(y_1^t) = \hat{\theta}^{t-1}(x)-\tau$ and $\theta(y_2^t) = \hat{\theta}^{t-1}(x)+\tau$. The range $[\theta(y_1^t), \theta(y_2^t)]$ is referred to as the search window, which is centered around the previous estimate $\hat{\theta}^{t-1}(x)$. Within the search window, the $\rho$-regressor regresses the $\rho$-rank $\hat{\rho}(x, y_{1}^{t}, y_{2}^{t})$, which is then converted to
\begin{eqnarray}
 \hat{\theta}^{t}(x) &=& \mbox{round}\big(\hat{\rho}(x, y_{1}^{t}, y_{2}^{t}) \cdot \tau(y_{1}^{t}, y_{2}^{t}) + \mu(y_{1}^{t}, y_{2}^{t})\big) \nonumber \\
    &=& \mbox{round}\big(\hat{\rho}(x, y_{1}^{t}, y_{2}^{t}) \cdot \tau + \hat{\theta}^{t-1}(x)\big).  \label{eq:relative_age_conversion}
\end{eqnarray}
The equality in \eqref{eq:relative_age_conversion} holds because $\tau$ is fixed and $\mu(y_{1}^{t}, y_{2}^{t}) =\hat{\theta}^{t-1}(x)$. This MWR is repeated until $\hat{\theta}^{t}(x) = \hat{\theta}^{t-1}(x)$ or a predefined number of iterations is reached. Note that the iteration terminates when the estimated rank is at the center of the search window, as shown in Iteration 4 in Figure~\ref{fig:MWM}(b).

\subsection{Global and Local Regression}
\label{ssec:GR_LR}
For accurate rank estimation, a network should be capable of learning various patterns. In facial age estimation, the facial aging process exhibits nonlinearity because each age group has different aging characteristics~\cite{albert2007age, zebrowitz1997reading}. During young ages, faces get bigger as they grow up. From middle to old ages, facial texture varies mainly due to skin aging. A global $\rho$-regressor, which is used for the entire age range, should learn these diverse aging patterns. On the other hand, a local $\rho$-regressor would be more effective for a specific age range if it were trained with images within that range only. The training also would be easier, because only the patterns in the smaller range should be learned~\cite{chen2017ranking}.

\begin{figure}[t]
    \centering
    \subfloat[]{\includegraphics[width=\linewidth]{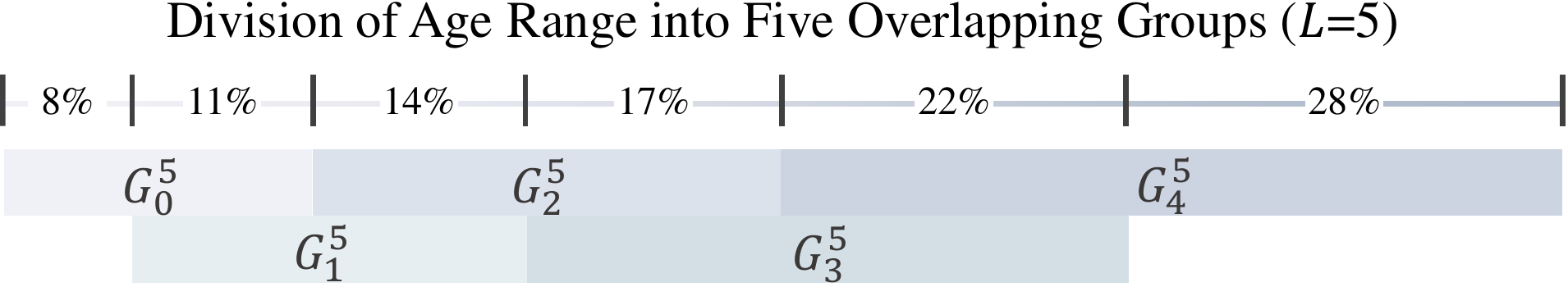}} \\
    \subfloat[]{\includegraphics[width=\linewidth]{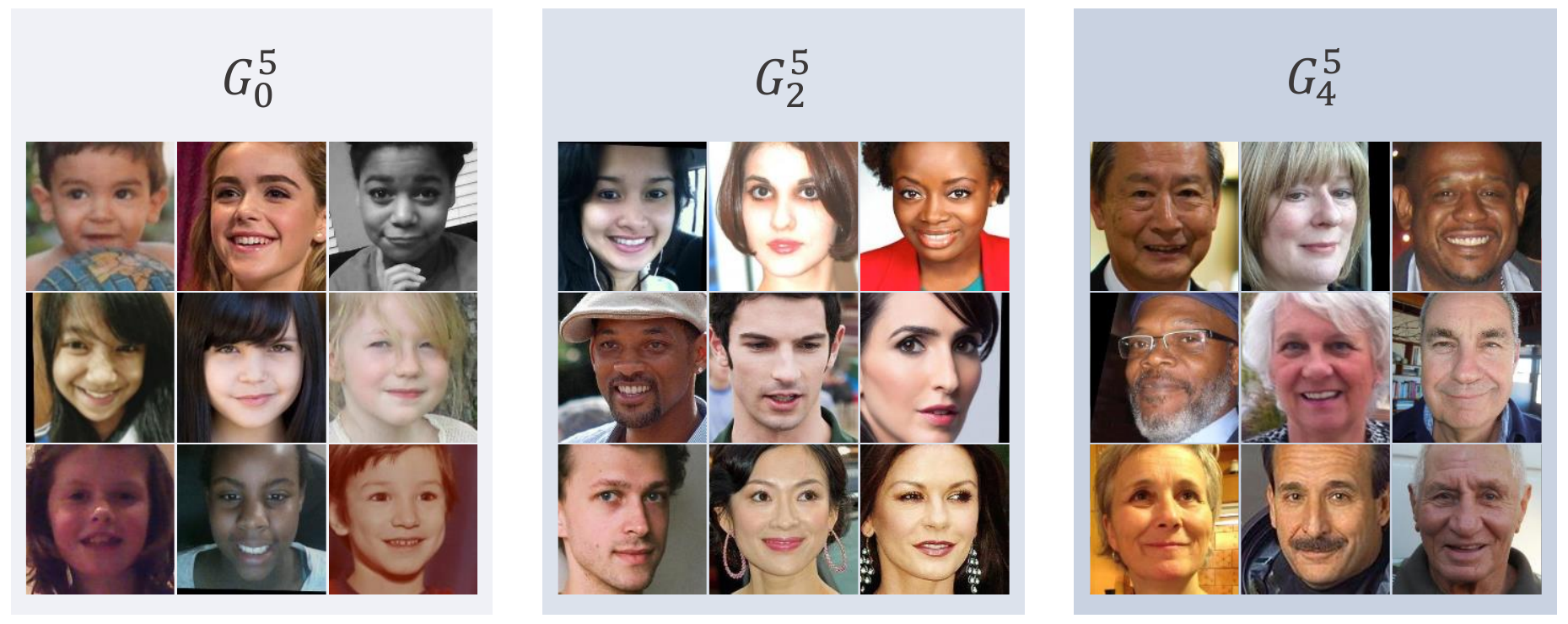}}
    \caption{An example of the age range partition: (a) The entire age range is divided into five overlapping groups $G_i^5$, $i=0,\ldots,4$. (b) Images are sampled from groups $G_0^5$, $G_2^5$, and $G_4^5$ in the CLAP2015 dataset \cite{escalera2015chalearn}.}
    \label{fig:3-5-partition}
\end{figure}

\begin{table*}[t]
    \caption
    {
        Comparison of facial age estimation results in the four evaluation settings (A, B, C, and D) of MORPH \RomNum{2} and on FG-NET.
    }
    \vspace*{-0.2cm}
    \centering
    \footnotesize
    \begin{tabular}{@{} l c c c c c c c c c c c c c c c c @{} }
        \toprule
        && \multicolumn{2}{c}{Setting A} && \multicolumn{2}{c}{Setting B} && \multicolumn{2}{c}{Setting C} && \multicolumn{2}{c}{Setting D} && \multicolumn{2}{c}{FG-NET} \\
        \cmidrule(lr){3-4} \cmidrule(lr){6-7} \cmidrule(lr){9-10} \cmidrule(lr){12-13} \cmidrule(lr){15-16}
        Algorithm && MAE & CS (\%) && MAE & CS (\%) && MAE & CS (\%) && MAE & CS (\%) && MAE & CS (\%) \\
        \midrule
          OR-CNN~\cite{niu2016ordinal} &&
          -    &  -   &&  -   &  -   &&
          -    &  -   && 3.27 & 73.0 && -  &  -  \\
          Ranking-CNN~\cite{chen2017ranking} &&
          -    &  -   &&  -   &  -   &&
          -    &  -   && 2.96 & 85.0 && -  &  -  \\
          DMTL~\cite{han2017heterogeneous} &&
          -    &  -   &&  -   &  -   &&
          3.00 & 85.3 &&  -   &  -   &&  - &  -  \\
          CMT~\cite{yoo2018deep} &&
          -    &  -   &&  -   &  -   &&
          2.91 & - &&  -   &  -   &&  - &  -  \\
          DEX~\cite{rothe2018imdb} &&
          2.68 &  -   &&  -   &  -  &&
          -    &  -   &&  -   &  -   &&  - &  -  \\
          DRFs~\cite{shen2018forests} &&
          2.91 & 82.9 && 2.98 &  -   &&
          -    &  - && 2.17 & 91.3   && 3.85  & 80.6   \\
          AGEn\cite{tan2018efficient} &&
          2.52 & 85.0 && 2.70 &  -   &&
            -  &  -   &&   -  &  -   &&  2.96 & 85.0  \\
          MV~\cite{pan2018mean} &&
          -    &  -   &&   -  &  -   &&
          2.79 &  -  && 2.16 &  -  && -  &  -  \\
          C3AE~\cite{zhang2019c3ae} &&
          -    &  -   &&   -  &  -   &&
          - &  -  && 2.75 &  -  && 2.95  &  -  \\
          BridgeNet~\cite{li2019bridgenet} &&
          2.38 & 91.0 && 2.63 & 86.0 &&
          -    &  -   &&   -  &  -   &&  2.56 &  \underline{86.0}  \\
          AVDL~\cite{wen2020adaptive} &&
          2.37 &  -   && \underline{2.53} &  - &&
          -    &  -  && \textbf{1.94}  &  - && \underline{2.32}  &  -  \\
          OL~\cite{lim2020order} &&
          2.41 & 91.7 && 2.75 & 88.2 &&
          2.68 & 88.8 && 2.22 & 93.3 && -  &  -  \\
          DRC-ORID~\cite{lee2021repulsive} &&
          \underline{2.26} & \underline{93.8} && \textbf{2.51}    & \underline{89.7} &&
          \underline{2.58} & \underline{89.5} && 2.16 & \underline{93.5} && -  &  -  \\
        \midrule
          Proposed &&
          \textbf{2.13} & \textbf{94.2} &&  \underline{2.53}   & \textbf{90.4} &&
          \textbf{2.53} & \textbf{90.5} &&   \underline{2.00}  & \textbf{95.0} && \textbf{2.23}  &  \textbf{91.1}  \\
        \bottomrule
    \end{tabular}
    \vspace*{-0.1cm}
   \label{table:morph_results}
\end{table*}

Hence, we employ a global $\rho$-regressor and $L$ local $\rho$-regressors. To use local $\rho$-regressors, we divide the entire rank range into multiple groups. For example, in facial age estimation, the configuration $L=5$ can be considered: First, the entire age range is partitioned into three age groups $G_0^5$, $G_2^5$, and $G_4^5$ in Figure~\ref{fig:3-5-partition}(a). Then, in addition to the three groups, we use two more groups $G_1^5$ and $G_3^5$ in Figure~\ref{fig:3-5-partition}(a). The lengths of the five overlapping groups form a geometric sequence, and $G_4^5$ covers the older half of the entire range. Thus, in the CLAP2015 dataset~\cite{escalera2015chalearn}, the entire range $[3, 85]$ is divided into five groups $[3, 18]$, $[10, 29]$, $[19, 44]$, $[30, 62]$, and $[45, 85]$. Figure~\ref{fig:3-5-partition}(b) shows some faces in group $G_0^5$, $G_2^5$, and $G_4^5$.

We can estimate the rank of an instance globally and then locally. First, we perform MWR using the global $\rho$-regressor to obtain the estimate $\hat{\theta}_{\rm global}(x)$. Second, starting with $\hat{\theta}_{\rm global}(x)$ for fast convergence, we iteratively refine the estimate using the $L$ local $\rho$-regressors to yield $\hat{\theta}_{\rm local}(x)$ eventually. At each iteration, due to the overlap, the previous estimate may belong to two rank groups. In such a case, both groups are selected and the estimated ranks from the corresponding local $\rho$-regressors are averaged.

To facilitate the transition between local $\rho$-regressors during the iterative MWR process, we train each local $\rho$-regressor using instances not only in the corresponding rank group but also those in nearby rank groups. More specifically, let $G_i = [\theta_{\min}, \theta_{\max}]$ denote the $i$th rank group, where $\theta_{\min}$ and $\theta_{\max}$ are the minimum and maximum ranks in the group. Then, the $i$-th local $\rho$-regressor is trained using instances $x$ whose ranks $\theta(x)$ are within an extended range $[\theta_{\min}-\alpha, \theta_{\max}+\alpha]$, where $\alpha$ is set to 6 for age estimation.

\subsection{Reference Selection}
\label{ssec:RS}
To predict the rank of a test instance, we compare it with reference pairs. We select these pairs from the training set offline prior to testing, based on the regression error $\gamma$, given by
\begin{equation}
\vspace{-0.12cm}
\gamma(y_1, y_2)=\frac{1}{|W|}\sum_{x \in W} |\hat{\rho}(x,y_1,y_2)-\rho(x,y_1,y_2) |
\vspace{-0.12cm}
\label{eq:ps}
\end{equation}
where $W=\{x \, | \, \theta(x) \in [\theta(y_1)-\alpha, \theta(y_2)+\alpha]\}$. The regression error $\gamma(y_1, y_2)$ represents the average estimation error of $\rho$-ranks, when $(y_1, y_2)$ is used as the reference pair. Thus, at iteration $t$ of the MWR process, we use the optimal pair $(y_1, y_2)$ with the smallest $\gamma(y_1, y_2)$, which satisfies the constraints of $\theta(y_1) = \hat{\theta}^{t-1}(x)-\tau$ and $\theta(y_2) = \hat{\theta}^{t-1}(x)+\tau$. Alternative reference selection schemes will be compared in Section~\ref{ssec:analysis}.

\section{Experimental Results}
\label{ssec:experiment}

We assess the performances of the proposed MWR on facial age estimation and HCI classification.

\subsection{Implementation Details} To train each $\rho$-regressor, we initialize its encoder using VGG16 pre-trained on ILSVRC2012 \cite{deng2009imagenet} and its regressor with the Kaiming uniform method \cite{he2015delving}. In facial age estimation, we do random horizontal flipping and random cropping to the image size of $224 \times 224$ for data augmentation. In HCI classification, we do random horizontal flipping only. We use the Adam optimizer \cite{kingma2014adam} with a learning rate of $10^{-4}$. The minibatch size is $18$. We use a PC with an Intel i7 processor and an NVIDIA RTX 2080Ti GPU.

\subsection{Facial Age Estimation}

\noindent\textbf{Datasets:} For facial age estimation, we use seven datasets: MORPH \RomNum{2} \cite{ricanek2006morph}, FG-NET \cite{lanitis2002toward}, CLAP2015 \cite{escalera2015chalearn}, UTK \cite{zhang2017age}, CACD \cite{chen2015face}, Adience \cite{levi2015age}, and IMDB-WIKI~\cite{rothe2018imdb}. We align all facial images, expect for Adience images, using landmarks detected by MTCNN \cite{zhang2016joint}, as done in \cite{li2019bridgenet}. For the Adience dataset, we use aligned images in \cite{levi2015age}. Unless specified otherwise, we pre-train the $\rho$-regressors on the IMDB-WIKI dataset, as done in \cite{rothe2018imdb, pan2018mean, tan2018efficient, li2019bridgenet, lim2020order, wen2020adaptive, lee2021repulsive}. Details about the datasets and experimental settings are in the supplemental document.

\vspace*{0.15cm}
\noindent\textbf{Geometric scheme:} Compared to young people, it is harder to estimate old people's ages; telling the difference between a 5-year-old and a 10-year-old is easier than that between a 65-year-old and a 70-year-old. Thus, for the relative age estimation of an old person, it is more effective to use a large search window. But, in \eqref{eq:arithmetic_tau}, the size of a search window $[\theta(y_1), \theta(y_2)]$ is fixed by $\tau$, regardless of the age $\theta(x)$ of input $x$. This is called an arithmetic $\tau$ and denoted by $\tau_{\rm ari}$, because the arithmetic difference is fixed. Instead, the geometric ratio between two reference ages can be fixed by
\begin{equation}
\tau_{\text{geo}} = \frac{1}{2}\big(\log \theta(y_2)-\log \theta(y_1) \big).
\label{eq:geo_tau}
\end{equation}
In this geometric scheme, the MWR process in~\eqref{eq:relative_age}$\sim$~\eqref{eq:relative_age_conversion} is modified by replacing each age $\theta(\cdot)$ with the logarithmic age $\log \theta(\cdot)$. Thus, a bigger search window is used for an older person. The default mode uses the geometric scheme with $\tau_{\rm geo}=0.1$. The arithmetic and geometric schemes will be compared in Section~\ref{ssec:analysis}.

\begin{table}[t]
    \caption
    {
        Comparison on the validation and test splits of CLAP2015.
    }
    \vspace*{-0.2cm}
    \centering
    \footnotesize
    \begin{tabular}{@{} l c c c c c c c c c  @{} }
        \toprule
        && \multicolumn{2}{c}{Validataion} && \multicolumn{2}{c}{Test} \\
        \cmidrule(lr){3-4} \cmidrule(lr){6-7}
        Algorithm && MAE & $\epsilon$-error && MAE & $\epsilon$-error  \\
        \midrule
        AgeNet~\cite{liu2015agenet}            && 3.33 & 0.29 && - & 0.26 \\
        Zhu \etal~\cite{zhu2015age}            && -    & 0.31 && - & 0.29 \\
        DEX~\cite{rothe2018imdb}               && 3.25 & 0.28 && - & 0.26 \\
        AGEn~\cite{tan2018efficient}         && 3.21 & 0.28 && 2.94 & 0.26 \\
        BridgeNet~\cite{li2019bridgenet}       && 2.98 & \textbf{0.26} && 2.87 & 0.26 \\

        \midrule
        Proposed                               && \textbf{2.95} & \textbf{0.26} && \textbf{2.77} & \textbf{0.25}  \\
        \bottomrule
    \end{tabular}
   \label{table:clap_results}
    \vspace*{-0.1cm}
\end{table}

\vspace*{0.15cm}
\noindent\textbf{Comparison with the state-of-the-arts:} Table~\ref{table:morph_results} compares the proposed algorithm with conventional algorithms in the four evaluation settings of MORPH~\RomNum{2} \cite{ricanek2006morph} and also on FG-NET \cite{lanitis2002toward}. We use the mean absolute error (MAE) and cumulative score (CS) metrics. MAE is the average absolute error between predicted and ground-truth ages, and CS is the percentage of images whose absolute errors are less than or equal to a tolerance level $l$. As in~\cite{lee2021repulsive, lim2020order, wen2020adaptive}, $l=5$. For the local regression, we set $L=5$, as shown in Figure~\ref{fig:3-5-partition}.

In Table~\ref{table:morph_results}, the proposed algorithm provides better results than the conventional algorithms in most tests. Compared with the state-of-the-art DRC-ORID~\cite{lee2021repulsive}, the proposed algorithm yields better CS scores in all four settings of MORPH \RomNum{2} and better MAE scores in three out of the four settings. The performance gaps are significant in many cases. For example, in setting D, the proposed algorithm reduces MAE by 7.4\% (from 2.16 to 2.00) and improves the CS score by 1.5\%. Similar to the proposed algorithm, the order learning methods in~\cite{lim2020order, lee2021repulsive} do relative comparisons. However, whereas these methods predict discrete order relations between instances, the proposed algorithm regresses continuous relative ages.

Table~\ref{table:morph_results} also compares the results on FG-NET \cite{lanitis2002toward}, which is a small dataset containing about 900 training images only in each fold. The proposed algorithm provides superior results by significant gaps of 0.09 in MAE and 5.1\% in CS. These excellent results on FG-NET confirm that the proposed algorithm learns aging characteristics effectively even from a small number of training images.

Next, Table~\ref{table:clap_results} compares the results on CLAP2015 \cite{escalera2015chalearn} using the metrics of MAE and $\epsilon$-error. For each image, CLAP2015 provides the standard deviation of age ratings by multiple annotators. The standard deviation represents the estimation difficulty. By taking this difficulty into account, $\epsilon$-error is defined as $1 - \text{exp}(-\frac{(\hat{\theta}(x)-\theta(x))^2}{2 \sigma^2})$, where $\sigma$ is the standard deviation of a test image $x$. The average $\epsilon$-error over all test images is reported. Both validation and test splits of CLAP2015 are used. For evaluation on the test set, we use the validation set, as well as the training set, to train the global and local $\rho$-regressors, as in~\cite{li2019bridgenet, tan2018efficient, liu2015agenet}.

The proposed algorithm achieves the best performances on both splits. CLAP2015 is a challenging dataset, so many conventional algorithms adopt performance boosting schemes. For example, BridgeNet~\cite{li2019bridgenet} improves its performance by averaging the predictions on 10 flipped and cropped images. AgeNet \cite{liu2015agenet}, DEX \cite{rothe2018imdb}, and AGEn \cite{tan2018efficient} employ eight or more networks. However, without using such schemes, the proposed algorithm outperforms all conventional algorithms. Especially, a significant MAE margin of $0.1$ is achieved on the test split.

\begin{table}[t]
    \caption
    {
        Comparison on UTK and in the train and validation settings of CACD. For both datasets, IMDB-WIKI pre-training is not performed.
    }
    \vspace*{-0.2cm}
    \centering
    \footnotesize
    \begin{tabular}{@{} l c c c c c c c c c c c c c c c @{} }
        \toprule
        &  UTK & Train & Validation \\
        \cmidrule(lr){2-2} \cmidrule(lr){3-3} \cmidrule(lr){4-4}
        Algorithm & MAE & MAE & MAE  \\
        \midrule
        dLDLF~\cite{shen2017label}       & - & 4.73 & 6.77 \\
        AGEn~\cite{tan2018efficient}   & - & 4.68 & - \\
        DRFs~\cite{shen2018forests}      & - & \underline{4.64} & \underline{5.77} \\
        CORAL~\cite{cao2019rank}         & 5.47 & - & -  \\
        Gustafsson \etal~\cite{gustafsson2020energy}        & 4.65 & - & -  \\
        Berg \etal~\cite{berg2021deep}   & \underline{4.55} & - & -  \\
        \midrule
        Proposed          & \textbf{4.37} & \textbf{4.41} & \textbf{5.68} \\
        \bottomrule
    \end{tabular}
   \label{table:utk_cacd_results}
\end{table}

\begin{table}[t]
    \caption
    {
        Accuracy and MAE comparison on the Adience and HCI datasets. For Adience, IMDB-WIKI pre-training is not performed.
    }
    \vspace*{-0.2cm}
    \centering
    \footnotesize
    \begin{tabular}{@{} l c c c c c c c@{} }
        \toprule
        & \multicolumn{2}{c}{Adience} & \multicolumn{2}{c}{HCI} \\
        \cmidrule(lr){2-3} \cmidrule(lr){4-5}
        Algorithm & Accuracy (\%) & MAE  & Accuracy (\%) & MAE\\
        \midrule
        Frank \& Hall~\cite{frank2001simple}   & -    & -    & 41.4 & 0.99     \\
        Cardoso \etal ~\cite{cardoso2007learning}   & -    & -    & 41.3 & 0.95     \\
        Palermo~\etal~\cite{palermo2012dating}   & -    & -    & 44.9 & 0.93     \\
        RED-SVM~\cite{lin2012reduction}          & -    & -    & 35.9 & 0.96     \\
        Martin~\etal~\cite{martin2014dating}     & -    & -    & 42.8 & 0.87     \\
        OR-CNN~\cite{niu2016ordinal}             & 56.7    & 0.54    & 38.7 & 0.95     \\
        CNNPOR~\cite{liu2018constrained}         & 57.4 & 0.55 & 50.1 & 0.82  \\
        GP-DNNOR~\cite{liu2019probabilistic}     & 57.4 & 0.54 & 46.6 & 0.76  \\
        SORD~\cite{diaz2019soft}                 & 59.6 & 0.49 & -    & -     \\
        DRC-ORID~\cite{lee2021repulsive}         & -    & -    & 44.7    & 0.80     \\
        Li~\etal~\cite{li2021ordinal}            & \underline{60.5} & \underline{0.47} & \underline{54.7} & \underline{0.66}  \\
        \midrule
        Proposed                                 & \textbf{62.6} & \textbf{0.45} & \textbf{57.8} & \textbf{0.58}   \\
        \bottomrule
    \end{tabular}
   \label{table:adience_hci}
     \vspace*{-0.1cm}
\end{table}

Table~\ref{table:utk_cacd_results} lists the performances on the UTK dataset \cite{zhang2017age}. The proposed algorithm outperforms the conventional algorithms with meaningful margins. Notice that Gustafsson \etal \cite{gustafsson2020energy} and Berg \etal \cite{berg2021deep} employ the deeper ResNet50~\cite{he2016deep} as their backbone networks, whereas the proposed $\rho$-reg\-ressors use VGG16. Nevertheless, the proposed algorithm improves the MAE score by more than 0.18.

Table~\ref{table:utk_cacd_results} also compares the results on CACD \cite{chen2015face}, which is a big dataset with many noisy data. The proposed algorithm also outperforms the conventional algorithms by 0.23 and 0.09 in MAE on the train and validation splits, respectively.

Lastly, Table~\ref{table:adience_hci} compares the results on Adience \cite{levi2015age}, which is used for age group estimation. In this test, we adopt the arithmetic scheme with $\tau_{\rm ari}=2$. Also, we use three local $\rho$-regressors that cover three equal parts of the rank range, respectively. The proposed algorithm outperforms the conventional algorithms by significant gaps of 2.1\% in accuracy and 0.02 in MAE.

\subsection{HCI Classification}

HCI \cite{palermo2012dating} is a dataset for determining the decade when a photograph was taken. It contains images from five decades 1930s $\sim$ 1970s. There are 265 images in each decade. As done in \cite{palermo2012dating, liu2018constrained, liu2019probabilistic, li2021ordinal}, we randomly split those 265 images into three subsets: 210 for training, 5 for validation, and 50 for testing. Then, the 10-fold cross-validation is performed. We use reference pairs satisfying $\tau_{\rm ari}\geq1$. For the local regression, three local $\rho$-regressors are employed that cover three equal parts of the rank range, respectively. Table~\ref{table:adience_hci} also compares the results on the HCI dataset. The proposed algorithm provides superior results by significant gaps of 3.1\% in accuracy and 0.08 in MAE.

%

\subsection{Analysis}
\label{ssec:analysis}

\begin{table}[!]
    \caption
    {
         Comparison of reference selection schemes on the test split of CLAP2015. MAE/$\epsilon$-error performances are compared. For the random scheme, the mean and standard deviation of MAE and $\epsilon$-error of 5 evaluation results are reported.
    }
    \vspace*{-0.2cm}
    \centering
    \footnotesize
    \begin{tabular}{@{} c c c c @{} }
        \toprule
          Random & Min $\gamma$ & Max $\gamma$ \\
         \midrule
        2.82$\pm$0.01/0.26$\pm$0.01 & \textbf{2.77}/\textbf{0.25} & 3.17/0.30 \\
        \bottomrule
    \end{tabular}
   \label{table:ref_sel}
\end{table}

\begin{table}
    \caption
    {
        Comparison of global and local $\rho$-regressors.
    }
    \vspace*{-0.2cm}
    \centering
    \footnotesize
    \hspace*{-0.1cm}
    \begin{tabular}{@{} l c c c c c @{} }
        \toprule
                             & MORPH~\RomNum{2} & CLAP2015  & UTK  \\
                             & (MAE/CS) & (MAE/$\epsilon$-error) & (MAE) \\
        \midrule
          Global $\rho$-regressor                  & 2.24/93.5          & 2.82/0.26   & 4.49 \\   
          Local $\rho$-regressors                 & \textbf{2.13}/\textbf{94.2}  & \textbf{2.77}/\textbf{0.25}   & \textbf{4.37} \\         
        \bottomrule
    \end{tabular}
   \label{table:global_vs_local}
\end{table}

\begin{figure}[t]
    \centering
    \subfloat[]{\includegraphics[width=0.48\linewidth]{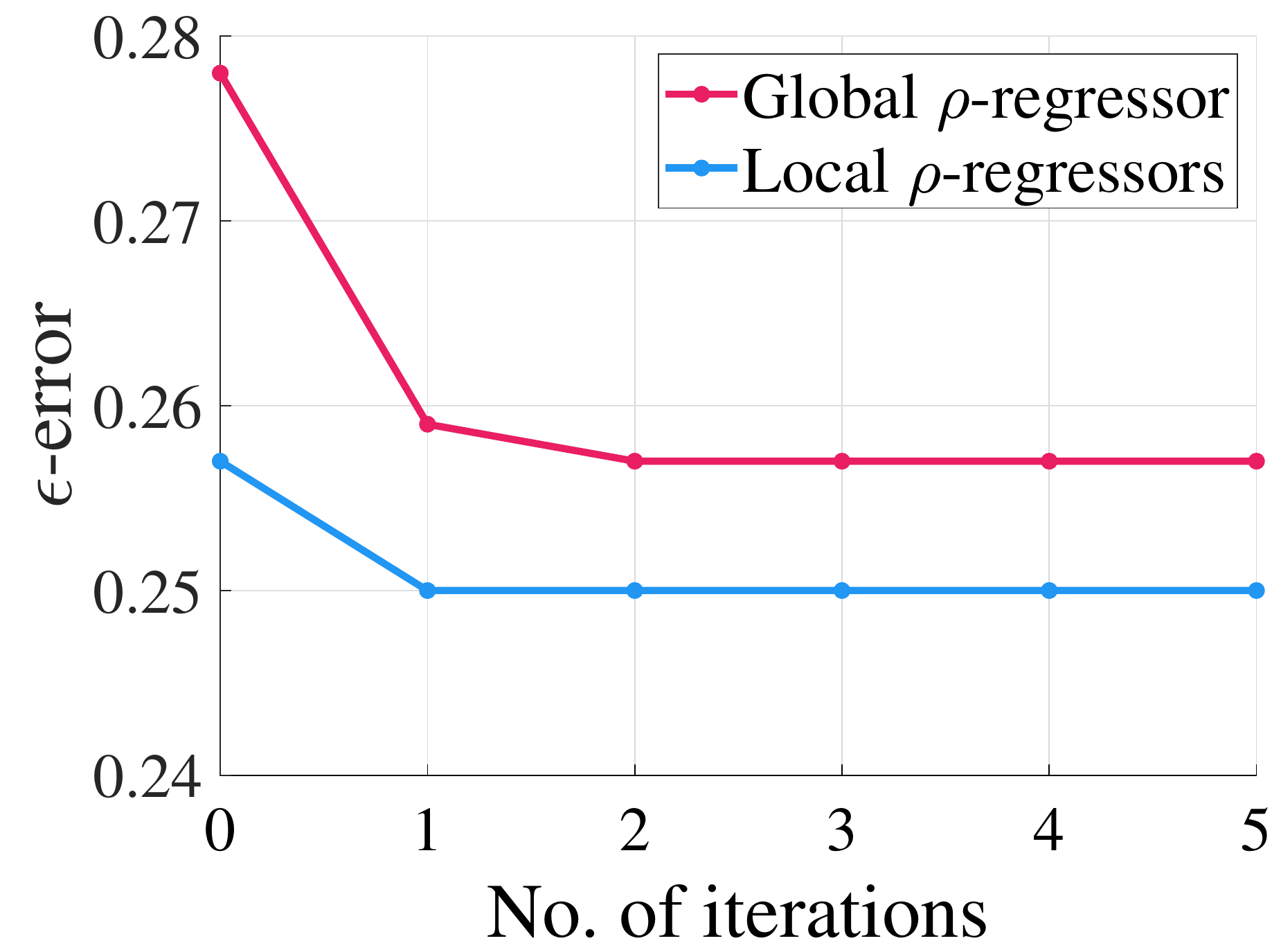}} \hspace{0.15cm}
    \subfloat[]{\includegraphics[width=0.48\linewidth]{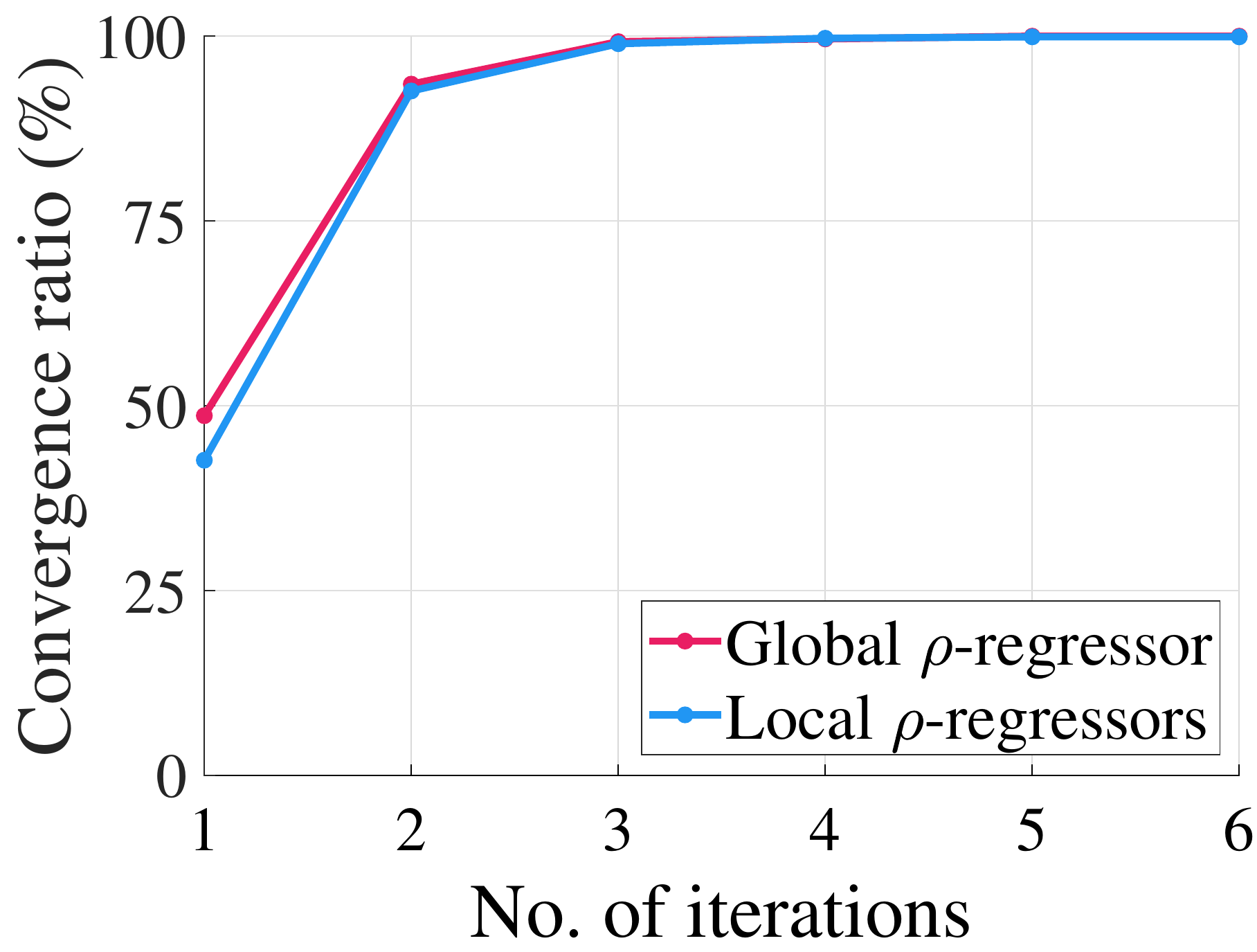}}
    \caption{Plots of (a) $\epsilon$-errors and (b) convergence ratios according to the number of iterations on the test split of CLAP2015. }
    \label{fig:iterations}
    \vspace*{-0.1cm}
\end{figure}

\begin{figure}[t]
    \centering
    \includegraphics[width=1\linewidth]{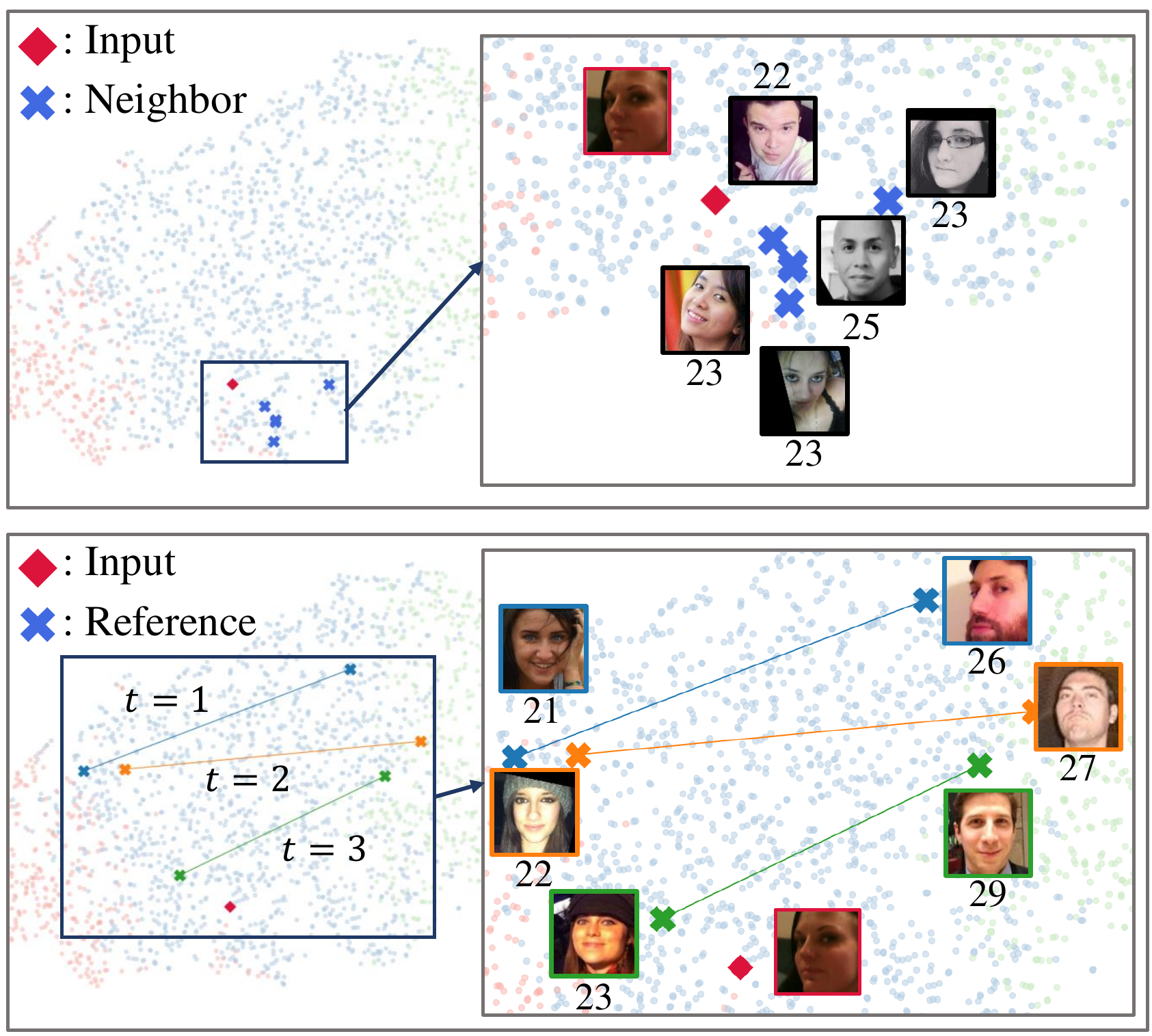}
        \caption{t-SNE visualization of MWR iterations. Top, the ground-truth age 27 of input $x$ is initially estimated to $\hat{\theta}^0(x)=23$ by finding the five NNs. Note that the 2D plot cannot perfectly preserve the order of all distances in the higher-dimensional feature space. Thus, the NNs in the feature space are not the nearest ones in this 2D plot. Above, the global estimate $\hat{\theta}_{\rm global}(x)=25$ is obtained in three iterations. The reference pair in each iteration is connected by an edge. In iterations 1, 2, and 3, the estimated ages are 24, 25, and 25, respectively.}
    \label{fig:result_example}
    \vspace*{-0.1cm}
\end{figure}

Next, we analyze the proposed MWR algorithm using facial age estimation datasets.

\vspace*{0.15cm}
\noindent\textbf{Reference selection:} Table~\ref{table:ref_sel} compares three reference selection schemes on CLAP2015. At iteration $t$ of the MWR process, a reference pair $(y_1, y_2)$ should satisfy the constraints of $\theta(y_1) = \hat{\theta}^{t-1}(x)-\tau$ and $\theta(y_2) = \hat{\theta}^{t-1}(x)+\tau$. Among all pairs satisfying the constraints, the random scheme selects one randomly. On the other hand, the min $\gamma$ or max $\gamma$ scheme chooses the pair with the smallest or largest regression error $\gamma$ in \eqref{eq:ps}, respectively. The min $\gamma$ scheme achieves the best results, so it is used as the default mode in this work. Examples of selected references are provided in the supplemental document.

\vspace*{0.15cm}
\noindent\textbf{Global vs. local regression:} Table~\ref{table:global_vs_local} compares the performances of global and local $\rho$-regressors in three test settings: MORPH~\RomNum{2} setting A, CLAP2015 test split, and UTK. First, only the global $\rho$-regressor is employed. In other words, MWR is performed without employing local $\rho$-regressors. Even the global regression outperforms the conventional state-of-the-art methods with meaningful margins in most cases. For example, on UTK, the global regression yields 0.06 lower MAE than the state-of-the-art Berg \etal \cite{berg2021deep}. Moreover, by employing local $\rho$-regressors, the proposed algorithm further improves the results. Especially, significant MAE improvements of 0.18 are achieved in comparison with Berg \etal \cite{berg2021deep} on UTK. More comparison results are presented in the supplemental document.

\begin{table}[t]
    \caption
    {
        Comparison of the arithmetic and geometric schemes on the test split of CLAP2015. Only the global $\rho$-regressor is used. The scores are slightly poorer than Table~\ref{table:clap_results}. This is because the training is performed for 60 epochs, while it is done for 130 epochs in Table~\ref{table:clap_results}.
    }
    \vspace*{-0.2cm}
    \centering
    \footnotesize
    \hspace*{-0.15cm}
    \begin{tabular}{@{} c c c c c c c c c@{} }
        \toprule
        & \multicolumn{3}{c}{$\tau_{\rm ari}$ for arithmetic scheme} & \multicolumn{4}{c}{$\tau_{\rm geo}$ for geometric scheme}\\
        \cmidrule(lr){2-4} \cmidrule(lr){5-8}
        & 3 & 5 & 7 & 0.05 & 0.075 & 0.1 & 0.125 \\
        \midrule
        MAE              & 3.21   & 3.33   & 3.24   & 3.03   & 2.95   & \textbf{2.93}   & 2.95   \\
        $\epsilon$-error & 0.304  & 0.311  & 0.303  & 0.276  & 0.271  & \textbf{0.266}  & 0.275 \\
        \bottomrule
    \end{tabular}
   \label{table:tau_study}
    \vspace*{-0.1cm}
\end{table}

\vspace*{0.15cm}
\noindent\textbf{Iterative MWR process:} To predict the rank of a test image, the proposed MWR iteratively compares it with references, which are selected from training images. The features of all references are extracted in advance for efficiency. Figure~\ref{fig:iterations}(a) plots how the age estimation performance varies as the MWR iteration goes on. In this test, we measure the $\epsilon$-errors on the test split of CLAP2015. In both global and local regression, the errors are reduced significantly at the first iterations and then saturate after the second iterations. Note that the local regression starts from global regression results $\hat{\theta}_{\rm global}(x)$.
Nevertheless, MWR further improves the results meaningfully using the local $\rho$-regressors. Figure~\ref{fig:iterations}(b) shows the ratio of cases that are converged up until each iteration. Although the convergence of MWR is not guaranteed theoretically, the global and local regression, respectively, converges within 6 iterations for most of images. In most cases, 4 iterations are sufficient. The result from the last iteration becomes the local regression estimate $\hat{\theta}_{\rm local}(x)$. Due to fast convergence, the total computing time for both global and local regression is only 0.007s (or equivalently 143fps) per image.


Figure~\ref{fig:result_example} visualizes an example of the MWR process in the embedding space, where the proposed algorithm obtains a global regression result $\hat{\theta}_{\rm global}(x)$ in three iterations.

\vspace*{0.15cm}
\noindent\textbf{Geometric vs. arithmetic schemes:} While the arithmetic scheme processes the entire age range identically, the geometric scheme in \eqref{eq:geo_tau} treats younger ages more finely using a smaller search range. Table~\ref{table:tau_study} compares these two schemes with various $\tau$'s on CLAP2015. In this test, only the global $\rho$-regressor is used. In general, the geometric scheme is better than the arithmetic scheme. The best performance is achieved by the geometric scheme with $\tau_\text{geo} = 0.1$, which is used in the default mode. The size of a search window varies in the geometric scheme. For example, at $\tau_\text{geo} = 0.1$, the size $\theta^{t}(y_2)-\theta^{t}(y_1)$ of the window is 3 for $\hat{\theta}^{t-1}(x) = 13$, while it is 12 for $\hat{\theta}^{t-1}(x) = 60$. In other words, a fine search is carried out near a young age estimate, while a coarse search is done near an old one.

\begin{figure}[t]
    \centering
    \subfloat[]{\includegraphics[width=1.0\linewidth]{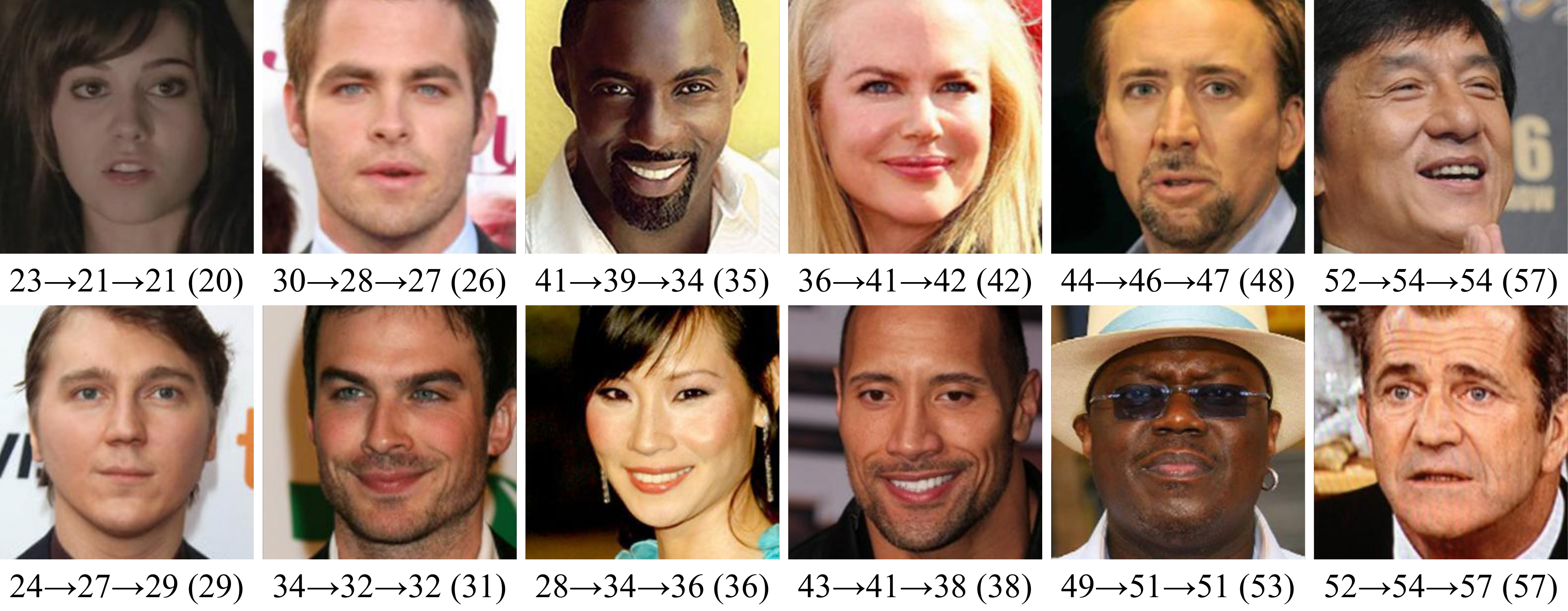}} \\ \vspace{0.05cm}
    \subfloat[]{\includegraphics[width=1.0\linewidth]{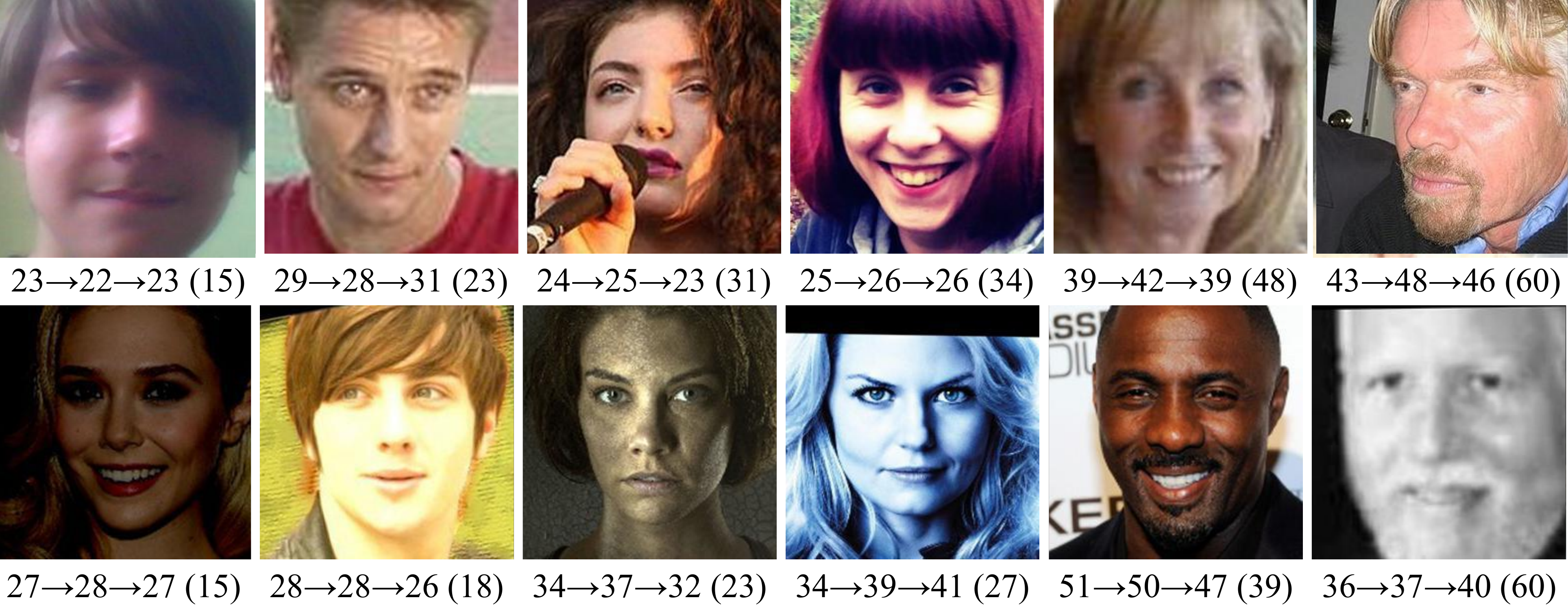}}
    \vspace{-0.1cm}
    \caption{(a) Success and (b) failure cases of the proposed algorithm in facial age estimation. For each image $x$, the initial, global, and local estimates $\hat{\theta}^{0}(x) \rightarrow \hat{\theta}_{\rm global}(x) \rightarrow \hat{\theta}_{\rm local}(x)$ are reported with the ground-truth $(\theta(x))$ within the parentheses.}
    \label{fig:MWR_examples}
    \vspace*{-0.1cm}
\end{figure}

\vspace*{0.15cm}
\noindent\textbf{Success and failure cases:} Figure~\ref{fig:MWR_examples} shows some success and failure cases of the proposed algorithm in facial age estimation. In (a), ages are estimated precisely with absolute errors less than 4. In (b), failure cases are shown, which are challenging examples due to various factors, such as low quality photographs, overexposure, and poor illumination.

\section{Conclusions}
We proposed a novel ordinal regression algorithm, called MWR. First, for relative ordinal regression, we designed global and local $\rho$-regressors. Then, we developed the MWR algorithm using these $\rho$-regressors. MWR first obtains an initial rank estimate based on the NN criterion. Then, it refines the estimate iteratively by selecting two reference instances to form a search window and estimating the $\rho$-rank within search window. Extensive experiments on various datasets showed that the proposed MWR algorithm provides outstanding rank estimation performances.

\section*{Acknowledgements}
\noindent This work was conducted partly by Center for Applied Research in Artificial Intelligence (CARAI) grant funded by DAPA and ADD (UD190031RD) and supported partly by the National Research Foundation of Korea (NRF) grants funded by the Korea government (MSIT) (No. NRF-2021R1A4A1031864 and  No. NRF-2022R1A2B5B03002310). 

\clearpage

{\small
\bibliographystyle{ieee_fullname}
\bibliography{cvpr2022_mwr}
}

\end{document}